\documentclass[final]{nesy2026} % Uncomment to include author names

\usepackage{longtable}% for long tables

\usepackage{booktabs}
\usepackage[load-configurations=version-1]{siunitx} % newer version
\usepackage{listings}
\usepackage{pgfplots}
\pgfplotsset{compat=1.18}

\usepackage{float}

\theorembodyfont{\upshape}
\theoremheaderfont{\scshape}
\theorempostheader{:}
\theoremsep{\newline}

\title[]{Text-to-SPARQL Generation with Reinforcement Learning: A GRPO-based Approach on DBLP}

\author{\Name{Jann Pfeifer} \Email{jann.pfeifer@leuphana.de}\\
\Name{Debayan Banerjee} \Email{debayan.banerjee@leuphana.de}\\
\Name{Ricardo Usbeck} \Email{ricardo.usbeck@leuphana.de}\\
\addr Leuphana University Lüneburg, Germany}

\begin{document}

\maketitle

\begin{abstract}
Knowledge graph question answering seeks to translate natural language questions into executable queries over knowledge graphs, but existing approaches often rely on large models or full supervision in the form of gold query annotations. 
This study examines whether reinforcement learning with outcome-based rewards can train a small instruction-tuned language model to perform zero-shot Text-to-SPARQL generation in the scholarly domain. 
Group Relative Policy Optimization (GRPO) is applied to the Qwen3-1.7B model on DBLP-QuAD, using prompts that combine natural language questions with symbolic hints about entities and relations. 
Training relies on execution feedback, structural constraints, and answer-level rewards, with an additional variant that incorporates gold-query-based shaping. 
The resulting models are compared to the unmodified zero-shot baseline and to a supervised DoRA-fine-tuned baseline across answer-level accuracy, execution accuracy, category-wise accuracy, and generalization to held-out templates. 
GRPO substantially improves over the zero-shot baseline and exhibits competitive generalization, while supervised DoRA fine-tuning achieves higher overall accuracy on the same model scale. 
Ablation analyses indicate that execution-based rewards account for most gains, with additional shaping yielding limited additional benefit, suggesting that outcome-based reinforcement learning is a viable training strategy when gold queries are unavailable for token-level supervision.
\end{abstract}

\section{Introduction}
Knowledge Graphs (KGs) provide a structured representation of entities and their relationships across domains such as healthcare, finance, and the scholarly literature \citep{abu2021domain}.
While SPARQL offers a flexible query language for such graphs, it requires users to understand the underlying schema and to adhere to its formal syntax, making direct interaction difficult for non-expert users.
Knowledge Graph Question Answering (KGQA) addresses this by mapping natural language questions to executable queries.

Early systems approached this as a semantic parsing task, training sequence models on datasets such as LC-QuAD \citep{trivedi2017lc} and QALD \citep{perevalov2022qald}.
Encoder--decoder architectures based on T5 and BART established strong supervised baselines under gold entity and relation linking \citep{banerjee2022modern}, with extensions including structure-aware pretraining \citep{qiEnhancingSPARQLQuery2024}, copy mechanisms for KG elements \citep{hirigoyenCopyMechanismHandling2022}, and enhanced attention with correction mechanisms \citep{chenEnhancingSPARQLQuery2025}.
More recently, LLMs have been applied via few-shot in-context learning and agentic pipelines \citep{dabramoInvestigatingLargeLanguage2025, strappazon2025instruct, avilaExperimentsTexttoSPARQLBased2024}, sometimes matching fine-tuned baselines without task-specific training.
However, these approaches typically rely on large models and substantial supervision or careful prompt engineering, which limits their applicability to new domains.

In the scholarly KGQA setting specifically, DBLP-QuAD \citep{banerjeeDBLPQuADQuestionAnswering2023} provides a 10,000-question benchmark over the DBLP bibliographic KG involving diverse question types, temporal conditions, and compositional patterns.
Other scholarly benchmarks include SciQA over the Open Research Knowledge Graph \citep{auer2023sciqa} and ORKG-QA \citep{jaradehQuestionAnsweringScholarly2020}.
Systems targeting these benchmarks range from neuro-symbolic pipelines \citep{abi2023psychic} to LLM prompting with ontological information \citep{jiang2023structure, meloniExploringLargeLanguage2025}, but they either rely on gold SPARQL supervision, assume strong linking components, or require large models.

Reinforcement Learning (RL) with verifiable rewards offers a complementary training paradigm for structured generation tasks where outputs can be executed and evaluated programmatically.
In the Text-to-SQL literature, RL has been applied with execution-based and constraint-based rewards \citep{chen2025constrainedsql,pourreza2025reasoningsql,stoisser2025sparks}, using algorithms such as REINFORCE, PPO, and Group Relative Policy Optimization (GRPO).
GRPO, introduced by \citet{shaoDeepSeekMathPushingLimits2024}, estimates advantages from groups of sampled trajectories without an explicit value function, making it suitable for memory-constrained RL fine-tuning of small models.
For KGQA, RL applications are more recent and typically agentic: \citet{vossebeldLearningRefineAgentic2025} fine-tune a compact LLM with GRPO to iteratively refine SPARQL over Wikidata, and concurrent preprints \citep{chen2025knowcoder, songEfficientTransferableAgentic2025a, zhang-zhao-2025-collaborative} adopt GRPO with curriculum or process-based rewards for multi-step KG reasoning.
However, these approaches mostly target iterative agent pipelines rather than single-step Text-to-SPARQL generation with small models.

A growing line of work shows that small models can be competitive for structured query generation when paired with appropriate training.
Parameter-efficient fine-tuning achieves competitive Text-to-SQL results at small scale \citep{karkiSmallerLargeLanguage2025}, RL fine-tuning can close part of the gap to large models on commodity hardware \citep{NGUYEN2025100135}, and sub-1B models have been applied to subgraph retrieval for KGQA \citep{huang2024less}.
\citet{jiang2025kg} further show that a 7B agentic model can outperform larger baselines.

Despite this progress, there is limited evidence on whether a small open-weight model can be trained via outcome-driven RL for single-step scholarly Text-to-SPARQL, and how such training compares to supervised fine-tuning at the same scale.
This work investigates this question by applying GRPO to Qwen3-1.7B \citep{yang2025qwen3technicalreport} on DBLP-QuAD.
The model receives the natural language question together with symbolic hints about relevant entities and relations, enriched with human-readable schema information, and produces a SPARQL query under a fixed output protocol.
We study two GRPO variants, one using only execution-based rewards and structural constraints, and one augmented with gold-query-based shaping. We compare both to the zero-shot base model and a supervised DoRA \citep{liu2024dora} baseline.

The contributions of this paper are threefold:
First, we present an outcome-driven training pipeline for scholarly Text-to-SPARQL that uses the symbolic executor as the primary supervision signal at small model scale.
Second, we provide a systematic comparison between zero-shot prompting, GRPO training, and supervised DoRA fine-tuning on DBLP-QuAD, including analyses of category-wise performance, temporal conditions, and held-out template generalization.
Third, we analyze multi-component reward design for GRPO in this setting, isolating the role of execution-based rewards relative to shaping terms.

\vspace{-2mm}
\section{Methodology}
\label{sec:methodology_paper}

\subsection{Task formulation and assumptions}
We study Text-to-SPARQL generation for scholarly KGQA on DBLP-QuAD, formulated as a single-step decision problem. Given one input prompt, the policy produces one complete SPARQL query as its terminal action. Inference is zero-shot, in the sense that no in-context demonstrations are provided at test time. 

A central design choice is to assume perfect entity and relation linking: the prompt provides pre-resolved entity and relation URIs alongside schema-enriched metadata, so the model is not required to perform entity disambiguation or relation mapping. 
This isolates the core challenge of generating an executable, KG-grounded query, while constituting a limitation by removing a major source of errors in end-to-end KGQA. 

To enable execution-based supervision, gold answer sets are materialized during preprocessing by executing the dataset's gold SPARQL queries against a fixed DBLP endpoint, and the resulting answer sets are used consistently for reward computation and evaluation.

\vspace{-2mm}
\subsection{Input representation as symbolic hints}
The policy input is a two-part conversational prompt consisting of a system message and a user message. This prompt functions as a lightweight symbolic interface, exposing relevant knowledge graph symbols (URIs) and typing cues (schema domain/range and short descriptions), and the neural policy is trained to respect these constraints during generation.

\vspace{-1mm}
\paragraph{System prompt.}
The system message specifies both the task and the required output protocol, enforcing the following constraints: the generated SPARQL query must be syntactically and semantically correct, use full URIs without prefixes, and default to \texttt{SELECT DISTINCT} for retrieval queries or \texttt{ASK} for yes/no questions. The model is further instructed to produce step-by-step internal reasoning enclosed within \texttt{<think>} tags, followed by the final SPARQL query only, with no additional explanation or formatting (\appendixref{app:system_prompt}).

\vspace{-1mm}
\paragraph{User prompt content.}
The user message contains the natural language question, a list of linked entities, and a list of linked relations. Each entity is provided as a URI together with a human-readable label, which allows the model to access the surface form without dereferencing URIs during inference. Each relation is provided as a URI, a label, and schema-level context (domain, range, and a short comment/description). These relation descriptors serve as a lightweight ontological grounding (\appendixref{app:user_prompt}).

\vspace{-3mm}
\subsection{GRPO training objective}
We optimize the policy using Group Relative Policy Optimization (GRPO). Intuitively, for each question we sample a small group of candidate queries, rank and normalize their rewards within the group, and update the policy to increase the likelihood of higher-reward queries while remaining close to a fixed reference policy via KL regularization.

Formally, for each prompt $x$ we sample $G$ rollouts $y_1,\dots,y_G \sim \pi_{\theta_{\text{old}}}(\cdot \mid x)$ and compute a scalar terminal reward $R(x,y_i)$ for each completion. Rewards are standardized within the group to obtain an implicit advantage estimate
\begin{align}
\hat{A}_i \;=\; \frac{R(x,y_i) - \mu_R}{\sigma_R + \varepsilon_{\text{std}}},
\end{align}
where $\mu_R$ and $\sigma_R$ denote the mean and standard deviation over the $G$ rewards. In all experiments, we use a fixed group size of $G=4$.

Policy updates follow a PPO-style clipped surrogate objective. Let
\begin{align}
r_i(\theta) \;=\; \frac{\pi_{\theta}(y_i \mid x)}{\pi_{\theta_{\text{old}}}(y_i \mid x)}
\end{align}
denote the probability ratio. The clipped objective is
\begin{align}
\mathcal{L}_{\text{clip}}(\theta)
= \mathbb{E}\Bigl[\min\bigl(r_i(\theta)\hat{A}_i,\; \mathrm{clip}(r_i(\theta), 1-\epsilon, 1+\epsilon)\hat{A}_i\bigr)\Bigr],
\end{align}
with clipping parameter $\epsilon = 0.2$. To regularize the updated policy against a fixed reference policy $\pi_{\text{ref}}$ (the base model prior to GRPO fine-tuning), we add a KL penalty and optimize
\begin{align}
\max_{\theta}\;\; \mathcal{L}_{\text{clip}}(\theta)\;-\;\beta \,\mathbb{E}\bigl[\mathrm{KL}(\pi_{\theta}(\cdot\mid x)\,\|\,\pi_{\text{ref}}(\cdot\mid x))\bigr],
\end{align}
with coefficient $\beta = 0.04$. Exploration is controlled through stochastic decoding and the KL constraint; no explicit entropy bonus is used.

\vspace{-3mm}
\subsection{Reward design}
The reward is designed to provide outcome-based supervision for executable query generation, with execution feedback as the dominant signal and several lower-weighted shaping terms that enforce symbolic well-formedness. For each completion, the final scalar reward is a heuristically weighted linear combination of six components:
\begin{align}
R \;=\; 3 R_{\text{exec}} + 2 R_{\text{sim}} + 1 R_{\text{struct}} + 0.5 R_{\text{format}} + 1 R_{\text{len}} + 1 R_{\text{len-ratio}}.
\end{align}

\vspace{-2mm}
\paragraph{Execution-feedback reward ($R_{\text{exec}}$).}
This component measures semantic correctness by executing the generated query $q_{\text{gen}}$ against the DBLP endpoint and comparing the answer set $A_{\text{gen}}$ to the gold answer set $A_{\text{gold}}$. If execution fails, the reward is set to a fixed penalty of $-0.5$. Otherwise, correctness is computed as the $F_1$ score over normalized answer tuples:
\vspace{-3mm}
\begin{align}
R_{\text{exec}} \;=\; F_1(A_{\text{gen}}, A_{\text{gold}}).
\end{align}

\vspace{-2mm}
\paragraph{Structural coverage reward ($R_{\text{struct}}$).}
To enforce grounding in the symbolic hints, we reward the inclusion of all required entity and relation URIs provided in the prompt:
\begin{align}
R_{\text{struct}} \;=\; 0.5\, I_{\text{relations}} \;+\; 0.5\, I_{\text{entities}},
\end{align}
where each indicator is $1$ if all required URIs of that type occur in $q_{\text{gen}}$ and $0$ otherwise. This discourages the model from ignoring parts of the provided context.

\vspace{-1mm}
\paragraph{Format reward ($R_{\text{format}}$).}
This component implements a soft output-protocol constraint. If a closing \texttt{</think>} tag is present, the suffix is extracted and checked for non-emptiness; a non-empty extracted query yields reward $1$, otherwise $0$. If no think tags are present, any non-empty completion is accepted with reward $1$.

\vspace{-1mm}
\paragraph{Length reward ($R_{\text{len}}$).}
To reduce truncation-related failures, we mildly penalize outputs that approach the maximum generation length. The reward decreases linearly once the token length $x$ exceeds a target length $n=768$ (maximum $a=1024$), clipped to $[0,1]$:
%\vspace{-2mm}
\begin{align}
R_{\text{len}} = \mathrm{clip}\left(1 - \frac{x - n}{a - n}, 0, 1\right).
\end{align}

\vspace{-3mm}
\paragraph{Gold-query shaping (optional).}
Two additional components are available only in settings where the gold SPARQL query is exposed during training. The query similarity reward $R_{\text{sim}}$ computes sentence-level BLEU between a normalized tokenization of $q_{\text{gen}}$ and the normalized gold query $q_{\text{gold}}$. The query length ratio reward $R_{\text{len-ratio}}$ regularizes query complexity via a symmetric log-distance between the token lengths of $q_{\text{gen}}$ and $q_{\text{gold}}$:
\vspace{-1mm}
\begin{align}
r = \frac{|q_{\text{gen}}|}{|q_{\text{gold}}|}, \qquad
R_{\text{len-ratio}} = \exp\bigl(-\alpha |\log r|\bigr), \quad \alpha=2.
\end{align}
These terms provide token-level shaping, but they also introduce competing objectives relative to execution-based correctness.

\vspace{-3mm}
\subsection{Post-processing and execution loop}
To robustly recover a final query candidate from raw completions, we locate the last closing \texttt{</think>} tag, discard all preceding content, and extract the last fenced code block from the remaining suffix if present, or use the full suffix otherwise. The extracted query is executed against a fixed Virtuoso endpoint hosting the April 2024 DBLP knowledge graph, with a normalized-query cache to avoid redundant endpoint calls during training and evaluation.

\vspace{-3mm}
\section{Experimental Setup}
\label{sec:exp_setup_paper}
\vspace{-1mm}
\subsection{Data and splits}
\label{sec:data_splits}

All experiments are conducted on DBLP-QuAD, a scholarly KGQA benchmark built on the DBLP knowledge graph. DBLP-QuAD contains 10{,}000 executable question--SPARQL pairs distributed via Hugging Face \citep{banerjeeDBLPQuADQuestionAnswering2023,banerjeeDBLPQuADQuestionAnswering2023Huggingface}, split into training, validation and test sets at a 7:1:2 ratio (7{,}000, 1{,}000, 
and 2{,}000 instances respectively).
Each instance provides the natural language question, the corresponding SPARQL query, linked entity and relation URIs, a query-type label, template identifier, and Boolean flags marking temporal questions and held-out template instances.

The query-type label contains one of ten semantic question categories: Single Fact, Multiple Facts, Boolean, Negation, Double Negation, Double Intent, Union, Count, Superlative/Comparative, and Disambiguation. Since these categories correspond to distinct logical query patterns, we report category-wise breakdowns in addition to aggregate scores.
We additionally evaluate on two orthogonal slices defined by the dataset flags. First, temporal questions, which require reasoning over publication years or other time-dependent information.
Second, held-out template instances, which account for roughly 19\% of the test split and provide a non-i.i.d. slice for assessing generalization beyond query patterns seen during training \citep{banerjeeDBLPQuADQuestionAnswering2023}.

\vspace{-2mm}
\subsection{Models and baselines}
\label{sec:models_baselines}

All compared configurations are based on the Qwen3-1.7B architecture and use the same preprocessing pipeline and prompt format described in Section~\ref{sec:methodology_paper}. 
% This yields a controlled comparison at constant model scale, where differences in performance can be attributed to the training objective and supervision regime.
A model from the Qwen3 family was chosen because it is practically trainable and has been shown to respond well to RL-style fine-tuning \citep{gandhi2025cognitive,wang2025octothinker,shao2025spurious}.

\vspace{-1mm}
\paragraph{Zero-shot base model.}
As a reference point, we evaluate the unmodified instruction-tuned base model in a zero-shot setting on DBLP-QuAD using the same prompt template as in all other configurations. We consider two prompting modes: a Chain-of-Thought (CoT) variant that requests intermediate reasoning, and a non-CoT variant that prompts the model to output the query directly. 
%In practice, the CoT variant frequently exceeds the maximum generation length for more complex questions, which yields truncated queries and a large fraction of execution failures. The non-CoT variant avoids this truncation effect and therefore provides a more conservative estimate of the base model's ability to produce valid SPARQL queries.

\vspace{-1mm}
\paragraph{Supervised fine-tuning baseline (DoRA).}
As a non-RL baseline, we train a supervised fine-tuned model using DoRA adapters on the gold SPARQL targets. We use DoRA as a parameter-efficient fine-tuning method primarily due to computational resource constraints. The model is optimized with a standard cross-entropy loss over the target query tokens, for a single epoch to match the data passes of the GRPO runs. Since the dataset provides supervision only for final SPARQL queries, CoT prompting is disabled in this setting.

\vspace{-1mm}
\paragraph{GRPO without gold-query shaping.}
Our main RL configuration uses GRPO with outcome-based rewards and does not assume access to gold SPARQL queries during training. Rewards are constructed from execution feedback and auxiliary shaping terms that enforce grounding in the provided entities and relations and stabilize training. This setting reflects a weakly supervised scenario where only questions and gold answers are available.

\vspace{-1mm}
\paragraph{GRPO with gold-query shaping.}
To assess the effect of incorporating query-level supervision within RL, we additionally evaluate a GRPO variant that includes gold-query-dependent shaping terms in the reward, namely a query similarity component and a query length regularizer. Comparing this variant to the outcome-only configuration isolates the contribution of gold-query shaping while keeping the RL algorithm and model scale fixed.

\vspace{-2mm}
\subsection{Training and compute conditions}
\label{sec:train_compute}

All runs are conducted on a single machine equipped with two NVIDIA RTX A6000 GPUs. Mixed-precision training with the bfloat16 format is enabled throughout all experiments. For GRPO, we fine-tune the full model parameters, whereas the supervised baseline is trained with DoRA adapters.

%To ensure a comparable evaluation regime across configurations, maximum input and output lengths are held constant. The combined system and user prompt is padded to a maximum of 668 tokens, which corresponds to the maximum tokenized prompt length encountered in the training set. Generated completions are limited to 1{,}024 tokens. The same prompt and completion limits are also applied when evaluating the zero-shot and supervised baselines.

GRPO training follows a fixed update scheme across the main configuration and all reward ablations. The per-device batch size is set to 4 questions, and for each question the policy samples a group of $G=4$ rollouts. This yields 16 sampled completions per forward pass on one device. Training is performed for a single epoch over the training set under the above accumulation regime. 
All runs use a fixed random seed for reproducibility. However, stochastic decoding during rollout generation (temperature 0.6, top-$p$ 0.95) introduces non-determinism that a fixed seed does not fully control.

Training is dominated by repeated SPARQL endpoint calls required for execution-feedback computation. On the described two-GPU setup, training the main GRPO configuration for one epoch requires approximately 30 hours. This runtime includes the overhead of endpoint calls during rollout evaluation, which remains the primary bottleneck despite query caching.

\vspace{-3mm}
\subsection{Metrics and evaluation protocol}
\label{sec:metrics_protocol}

Evaluation targets semantic correctness via execution. For each model, generated queries are executed against the fixed DBLP endpoint, and the resulting answer bindings are compared to the gold answer sets materialized during preprocessing. All metrics are computed on the held-out test split of 2{,}000 examples.

\vspace{-1mm}
\paragraph{Execution accuracy (ExAcc).}
Execution accuracy measures the fraction of instances for which the extracted query executes without raising an error. This metric captures adherence to SPARQL syntax and to the required output protocol.

\vspace{-1mm}
\paragraph{Answer exact match (EMAcc) and answer-set $F_1$.}
Exact match accuracy (EMAcc) records the fraction of instances where the executed answer set matches the gold answer set exactly. To capture partial correctness for multi-answer queries, we additionally compute an $F_1$ score over answer sets.

\vspace{-1mm}
\paragraph{Category-wise analysis.}
To analyze performance across distinct logical query patterns, we report category-wise exact-match accuracies for the ten DBLP-QuAD question categories.

\vspace{-1mm}
\paragraph{Temporal and held-out template slices.}
Temporal accuracy (TempAcc) is computed analogously to EMAcc but restricted to the subset of instances marked as temporal. To assess generalization to unseen forms, we report held-out template accuracy (GenAcc), computed as EMAcc on the instances derived from templates that are withheld from training.

\vspace{-2mm}
\section{Results}
\label{sec:results}
\subsection{Overall answer-level accuracy}
\label{sec:results:overall}

\begin{table}[t]
\floatconts
  {tab:answer-level-main}
  {\caption[Overall answer-level performance on DBLP-QuAD]{Overall answer-level performance on DBLP-QuAD. EMAcc denotes exact match accuracy on answer sets, ExAcc the fraction of queries that execute without errors, and F1 the macro-averaged F1 score over result sets.}}
  {%
  \vspace{-4mm}
  \begin{tabular}{lccc}
    \toprule
    \bfseries Model & \bfseries EMAcc & \bfseries ExAcc & \bfseries F1 \\
    \midrule
    Qwen3 1.7B Base & 0.07 & 0.50 & 0.14 \\
    Qwen3 1.7B Base (with CoT) & 0.11 & 0.23 & 0.11 \\
    \midrule
    Qwen3 1.7B DoRA (1 Epoch) & \textbf{0.69} & \textbf{0.91} & \textbf{0.78} \\
    \midrule
    Qwen3 1.7B GRPO (without gold queries) & \underline{0.47} & 0.83 & \underline{0.48} \\
    Qwen3 1.7B GRPO (with gold queries) & 0.44 & \underline{0.85} & 0.45 \\
    \bottomrule
  \end{tabular}
  }
\end{table}

Table~\ref{tab:answer-level-main} summarizes overall test set results. The zero-shot base model attains low answer-level performance (EMAcc 0.07, F1 0.14). Enabling chain-of-thought prompting increases EMAcc to 0.11, but sharply reduces executability (ExAcc 0.23) and slightly decreases F1 to 0.11. This pattern is consistent with CoT completions that deviate from the required output protocol or are truncated before producing a valid query.

The supervised DoRA baseline achieves the strongest results overall (EMAcc 0.69, ExAcc 0.91, F1 0.78). Both GRPO variants substantially improve over zero-shot prompting. GRPO without gold-query-based rewards reaches EMAcc 0.47, ExAcc 0.83, and F1 0.48. Adding gold-query-based reward terms increases ExAcc slightly to 0.85, but reduces EMAcc (0.44) and F1 (0.45). Overall, GRPO closes a considerable part of the gap between zero-shot prompting and supervised fine-tuning, while remaining clearly below DoRA on aggregate accuracy in this configuration.

\vspace{-3mm}
\subsection{Category-wise behavior}
\label{sec:results:categories}

Table~\ref{tab:category-acc} shows exact match accuracies across DBLP-QuAD question types. Categories corresponding to relatively local graph patterns (e.g., Single-Fact and Disambiguation) saturate at high scores for both DoRA and GRPO, suggesting that both methods reliably teach schema grounding and triple-pattern assembly. In contrast, disjunction (UNION) and Superlative/Comparative questions remain challenging, especially for GRPO, and constitute recurring failure modes in the qualitative analysis.

\begin{table}[t]
\floatconts
  {tab:category-acc}
  {\caption[Accuracy by question category]{Accuracy by question category.
  \emph{Bool} = Boolean, \emph{Cnt} = Count, \emph{SF} = Single-Fact, \emph{Un} = Union, etc.
  \emph{Base} = Qwen3-1.7B; \emph{DoRA} = Qwen3-1.7B with DoRA adapters;
  \emph{GRPO-no} / \emph{GRPO-gold} = GRPO without/with gold-query rewards.}}
  {%
  \vspace{-4mm}
  {\small
  \setlength{\tabcolsep}{3pt}% local to this table
  \begin{tabular*}{\linewidth}{@{\extracolsep{\fill}} l*{10}{c}}
    \toprule
    \bfseries Model & \bfseries Bool & \bfseries Cnt & \bfseries Disamb & \bfseries D-Int & \bfseries D-Neg & \bfseries Multi & \bfseries Neg & \bfseries SF & \bfseries Sup+Comp & \bfseries Un \\
    \midrule
    Base     & 0.00 & 0.10 & 0.07 & 0.04 & 0.00 & 0.09 & 0.00 & 0.25 & 0.14 & 0.04 \\
    Base+CoT & 0.16 & 0.05 & 0.08 & 0.08 & 0.06 & 0.10 & 0.12 & 0.44 & 0.00 & 0.01 \\
    \midrule
    DoRA     & \textbf{0.72} & \textbf{0.59} & \textbf{0.85} & \textbf{0.58} & \textbf{0.86} & \textbf{0.71} & \textbf{0.75} & \textbf{0.96} & \textbf{0.46} & \textbf{0.56} \\
    \midrule
    GRPO-no   & \underline{0.47} & \underline{0.40} & \underline{0.70} & 0.37 & \underline{0.63} & 0.40 & \underline{0.47} & \underline{0.93} & \underline{0.19} & \underline{0.27} \\
    GRPO-gold & 0.44 & 0.39 & 0.61 & \underline{0.40} & 0.60 & \underline{0.45} & 0.42 & 0.90 & 0.18 & 0.15 \\
    \bottomrule
  \end{tabular*}
  }% \small
  }
\end{table}

\vspace{-3mm}
\subsection{Generalization and temporal subsets}
\label{sec:results:slices}

Table~\ref{tab:temporal-generalization} reports exact match on the subsets for questions with temporal information and questions from held-out templates. For temporal questions, the ordering mirrors the overall results. DoRA performs best (TempAcc 0.69), followed by the GRPO variants (0.52 and 0.47), while the base models remain near 0.09--0.10. Thus, GRPO substantially improves over zero-shot prompting, yet does not close the gap to supervised fine-tuning.

%\vspace{-2mm}
%\vspace{-1mm}

On held-out templates, GRPO achieves the strongest performance. GRPO without gold-query shaping attains GenAcc 0.53 (and 0.48 with gold-query-based rewards), whereas DoRA reaches 0.40. Notably, GRPO generalization is close to its overall performance, while DoRA exhibits a larger drop from overall to held-out template accuracy.

\begin{table}[t]
\floatconts
  {tab:temporal-generalization}
  {\caption[Performance on temporal and held-out template questions]{Answer-level performance on temporal questions (TempAcc) and on questions derived from held-out templates (GenAcc), reported as exact match accuracy.}}
  {%
  \vspace{-4mm}
  \begin{tabular}{lcc}
    \toprule
    \bfseries Model & \bfseries TempAcc & \bfseries GenAcc \\
    \midrule
    Qwen3 1.7B Base & 0.09 & 0.06 \\
    Qwen3 1.7B Base (with CoT) & 0.10 & 0.15 \\
    \midrule
    Qwen3 1.7B DoRA (1 Epoch) & \textbf{0.69} & 0.40 \\
    \midrule
    Qwen3 1.7B GRPO (without gold queries) & \underline{0.52} & \textbf{0.53} \\
    Qwen3 1.7B GRPO (with gold queries) & 0.47 & \underline{0.48} \\
    \bottomrule
  \end{tabular}
  }
\end{table}

\vspace{-1mm}
\subsection{Reward ablations and error analysis}
\label{sec:results:ablations}
\vspace{-1mm}
\begin{table}[t]
\floatconts
  {tab:reward-ablations}
  {\caption[GRPO reward ablations]{Answer-level, execution, temporal, and generalization performance for GRPO reward ablations.
  Configurations differ only in the included reward components: execution feedback, format reward, structural coverage reward, length reward, and gold-query-based rewards.}}
  {%
  \vspace{-4mm}
  \setlength{\tabcolsep}{5pt}% local to this table
  \begin{tabular}{p{0.40\textwidth}ccccc}
    \toprule
    \bfseries Model & \bfseries EMAcc & \bfseries ExAcc & \bfseries F1 & \bfseries TempAcc & \bfseries GenAcc \\
    \midrule
    $R_{\text{exec}}$                                   & 0.40          & 0.71          & 0.41               & 0.40          & 0.49 \\
    $R_{\text{exec,format}}$                            & 0.42          & 0.80          & 0.43               & 0.41          & 0.49 \\
    $R_{\text{exec,format,struct}}$                     & \underline{0.46} & 0.80       & \underline{0.48}   & \underline{0.51} & \underline{0.51} \\
    $R_{\text{exec,format,struct,len}}$                 & \textbf{0.47} & \underline{0.83} & \underline{0.48} & \textbf{0.52} & \textbf{0.53} \\
    $R_{\text{exec,format,struct,len,gold(sim,len-r)}}$ & 0.44          & \textbf{0.85} & 0.45               & 0.47          & 0.48 \\
    \bottomrule
  \end{tabular}
  }
\end{table}

To isolate which reward components drive the GRPO gains, Table~\ref{tab:reward-ablations} reports ablations that differ only in reward composition. Execution-feedback already yields most of the final performance (EMAcc 0.40, F1 0.41). Adding a format reward yields modest answer-level gains but improves execution reliability (ExAcc 0.80). Introducing the structural coverage reward yields the most pronounced improvements in answer correctness and subset performance (F1 0.48, TempAcc/GenAcc 0.51). The length reward refines this configuration slightly, corresponding to the strongest no-gold setting (EMAcc 0.47, ExAcc 0.83, TempAcc 0.52, GenAcc 0.53).

Adding gold-query-based similarity and length-ratio terms increases ExAcc further (0.85) but reduces answer-level accuracy and subset performance. In this setup, gold-based shaping improves executability marginally while slightly degrading correctness and generalization.

A qualitative analysis of stratified samples of incorrect generations (100 for GRPO-no-gold; 100 for DoRA) shows that failures concentrate around a small set of symbolic operators and structural requirements: disjunction/UNION handling and answer-shape consistency, negation via \texttt{FILTER NOT EXISTS}, aggregation and superlatives, multi-intent joins and missing constraints, temporal comparisons, and syntax or schema violations.

The error profile differs by training strategy. GRPO exhibits more malformed outputs and structural errors (including syntax and schema issues that prevent execution), alongside frequent failures on UNION, negation, and multi-intent patterns. DoRA failures are more often subtle semantic mismatches within otherwise well-formed SPARQL queries, and its larger drop on held-out templates is consistent with increased template reliance. Both models frequently mis-handle Boolean question shapes (ASK vs.\ SELECT), producing answer-set mismatches even when partial patterns are correct. 
% Detailed error distributions and representative examples are provided in the supplementary material.

\vspace{-2mm}
\section{Discussion and Conclusion}
\label{sec:discussion_conclusion}
The results support a limited but practically relevant conclusion about outcome-based reinforcement learning for single-step Text-to-SPARQL at small model scale. In this setting, GRPO can improve a weak zero-shot baseline under a strict output protocol, but it does not substitute supervised fine-tuning when gold SPARQL targets are available. DoRA achieves stronger overall performance, which is consistent with the denser supervision signal from token-level cross-entropy on gold queries, whereas GRPO learns from a terminal reward assignment over long generations. GRPO is therefore best viewed as a viable alternative for weakly supervised conditions where direct query supervision is unavailable. 

A notable pattern concerns generalization on held-out templates. GRPO appears less sensitive to the template shift than DoRA supervised fine-tuning, which is consistent with outcome-based optimization rewarding semantic correctness independently of the exact surface form of a query. However, this observation should be interpreted cautiously, as the held-out slice is still derived from the same benchmark and shares schema, vocabulary, and construction procedure with the training data. As a result, it does not fully represent out-of-distribution generalization in the open-world sense, and further validation on additional datasets and query structures is required. 

The category-wise and qualitative analyses suggest that errors shift from grounding to composition as query complexity increases. For simpler patterns, outcome-based reinforcement learning appears sufficient to learn basic SPARQL assembly and to ground generations in the provided entities and schema hints. For more compositional queries, errors more often involve symbolic operators such as disjunction, aggregation, and multi-intent joins. This is consistent with sparse terminal feedback providing limited guidance for precise structural decisions, and it motivates reward designs that more directly target compositional structure.

The reward ablations support this interpretation. Execution feedback accounts for most GRPO gains, while structural coverage provides the most meaningful additional contribution by encouraging consistent grounding in the supplied schema information. In contrast, gold-query-based shaping does not improve answer-level performance in this setup, and it may introduce a competing signal that favors reproducing a particular reference surface form even when alternative formulations are semantically correct and executable.

Several limitations constrain the scope of these conclusions. The study assumes perfect entity and relation linking, which isolates SPARQL generation but removes a major failure source in end-to-end KGQA. The evaluation further focuses on a single model family, a single benchmark, and a fixed execution environment, so the observed patterns are not yet validated across knowledge graphs, schemas, or domains. Finally, results are based on single runs, and stochastic decoding during rollout generation introduces non-determinism that a fixed seed does not fully control. More extensive replication was limited by training time, since reward computation requires repeated SPARQL endpoint executions during rollout evaluation and this dominates runtime despite caching. Future work should integrate entity and relation linking into training and investigate process-level rewards derived from query decomposition to address failures on compositional query structures, potentially combined with curriculum strategies.

\acks{This work is supported by the German Research Foundation (DFG) project NFDI4DS under Grant No.: 460234259.}

\bibliography{nesy2026-sample}

@inproceedings{abi2023psychic,
  title = {{{PSYCHIC}}: A Neuro-Symbolic Framework for Knowledge Graph Question-Answering Grounding},
  booktitle = {{{ISWC}} 2023-{{International}} Semantic Web Conference},
  author = {Abi Akl, Hanna},
  year = 2023
}

@article{abu2021domain,
  title = {Domain-Specific Knowledge Graphs: {{A}} Survey},
  author = {{Abu-Salih}, Bilal},
  year = 2021,
  journal = {Journal of Network and Computer Applications},
  volume = {185},
  pages = {103076},
  publisher = {Elsevier}
}

@article{auer2023sciqa,
  title = {The Sciqa Scientific Question Answering Benchmark for Scholarly Knowledge},
  author = {Auer, S{\"o}ren and Barone, Dante AC and Bartz, Cassiano and Cortes, Eduardo G and Jaradeh, Mohamad Yaser and Karras, Oliver and Koubarakis, Manolis and Mouromtsev, Dmitry and Pliukhin, Dmitrii and Radyush, Daniil and others},
  year = 2023,
  journal = {Scientific Reports},
  volume = {13},
  number = {1},
  pages = {7240},
  publisher = {Nature Publishing Group UK London}
}

@inproceedings{avilaExperimentsTexttoSPARQLBased2024,
  title = {Experiments with Text-to-{{SPARQL}} Based on {{ChatGPT}}},
  booktitle = {2024 {{IEEE}} 18th International Conference on Semantic Computing ({{ICSC}})},
  author = {Avila, Caio Viktor S. and Vidal, V{\^a}nia M.P. and Franco, Wellington and Casanova, Marco A.},
  year = 2024,
  pages = {277--284},
  doi = {10.1109/ICSC59802.2024.00050}
}

@inproceedings{banerjee2022modern,
  title = {Modern Baselines for {{SPARQL}} Semantic Parsing},
  booktitle = {Proceedings of the 45th International {{ACM SIGIR}} Conference on Research and Development in Information Retrieval},
  author = {Banerjee, Debayan and Nair, Pranav Ajit and Kaur, Jivat Neet and Usbeck, Ricardo and Biemann, Chris},
  year = 2022,
  pages = {2260--2265}
}

@inproceedings{banerjeeDBLPQuADQuestionAnswering2023,
  title = {{{DBLP-QuAD}}: A Question Answering Dataset over the {{DBLP}} Scholarly Knowledge Graph},
  booktitle = {{{BIR}}@{{ECIR}}},
  author = {Banerjee, Debayan and Awale, Sushil and Usbeck, Ricardo and Biemann, Chris},
  year = 2023,
  pages = {37--51},
  cdate = {1672531200000}
}

@misc{banerjeeDBLPQuADQuestionAnswering2023Huggingface,
  title = {Awalesushil/{{DBLP-QuAD}}},
  author = {Banerjee, Debayan and Awale, Sushil and Usbeck, Ricardo and Biemann, Chris},
  year = 2023,
  publisher = {awalesushil/DBLP-QuAD},
  address = {https://huggingface.co/datasets/awalesushil/DBLP-QuAD}
}

@inproceedings{chen2025constrainedsql,
  title = {{{ConstrainedSQL}}: {{Training}} Llms for {{Text2SQL}} via Constrained Reinforcement Learning},
  booktitle = {Annual Conference on Neural Information Processing Systems},
  author = {Chen, Weiqin and Pham, Nhan and Glass, Michael and Vu, Long and Rossiello, Gaetano and Subramaniam, Shankar and Paternain, Santiago},
  year = 2025
}

@article{chen2025knowcoder,
  title = {{{KnowCoder-A1}}: {{Incentivizing}} Agentic Reasoning Capability with Outcome Supervision for {{KBQA}}},
  author = {Chen, Zhuo and Wang, Fei and Li, Zixuan and Zhang, Zhao and Ding, Weiwei and Yang, Chuanguang and Xu, Yongjun and Jin, Xiaolong and Guo, Jiafeng},
  year = 2025,
  journal = {arXiv preprint arXiv:2510.25101},
  eprint = {2510.25101},
  archiveprefix = {arXiv}
}

@article{chenEnhancingSPARQLQuery2025,
  title = {Enhancing {{SPARQL}} Query Generation for Question Answering with a Hybrid Encoder--Decoder and Cross-Attention Model},
  author = {Chen, Yi-Hui and Lu, Eric Jui-Lin and Cheng, Kwan-Ho},
  year = 2025,
  journal = {Journal of Web Semantics},
  volume = {87},
  pages = {100869},
  issn = {1570-8268},
  doi = {10.1016/j.websem.2025.100869}
}

@inproceedings{dabramoInvestigatingLargeLanguage2025,
  title = {Investigating {{Large Language Models}} for {{Text-to-SPARQL Generation}}},
  booktitle = {Proceedings of the 4th {{International Workshop}} on {{Knowledge-Augmented Methods}} for {{Natural Language Processing}}},
  author = {D'Abramo, Jacopo and Zugarini, Andrea and Torroni, Paolo},
  editor = {Shi, Weijia and Yu, Wenhao and Asai, Akari and Jiang, Meng and Durrett, Greg and Hajishirzi, Hannaneh and Zettlemoyer, Luke},
  year = 2025,
  month = may,
  pages = {66--80},
  publisher = {Association for Computational Linguistics},
  address = {Albuquerque, New Mexico, USA},
  urldate = {2025-05-25},
  isbn = {979-8-89176-229-9}
}

@inproceedings{gandhi2025cognitive,
  title = {Cognitive Behaviors That Enable Self-Improving Reasoners, or, Four Habits of Highly Effective {{STaRs}}},
  booktitle = {Second Conference on Language Modeling},
  author = {Gandhi, Kanishk and Chakravarthy, Ayush K and Singh, Anikait and Lile, Nathan and Goodman, Noah},
  year = 2025
}

@inproceedings{hirigoyenCopyMechanismHandling2022,
  title = {A Copy Mechanism for Handling Knowledge Base Elements in {{SPARQL}} Neural Machine Translation},
  booktitle = {Findings of the Association for Computational Linguistics: {{AACL-IJCNLP}} 2022},
  author = {Hirigoyen, Rose and Zouaq, Amal and Reyd, Samuel},
  editor = {He, Yulan and Ji, Heng and Li, Sujian and Liu, Yang and Chang, Chua-Hui},
  year = 2022,
  month = nov,
  pages = {226--236},
  publisher = {Association for Computational Linguistics},
  address = {Online only},
  doi = {10.18653/v1/2022.findings-aacl.22}
}

@inproceedings{huang2024less,
  title = {Less Is More: {{Making}} Smaller Language Models Competent Subgraph Retrievers for Multi-Hop {{KGQA}}},
  booktitle = {Findings of the Association for Computational Linguistics: {{EMNLP}} 2024},
  author = {Huang, Wenyu and Zhou, Guancheng and Wang, Hongru and Vougiouklis, Pavlos and Lapata, Mirella and Pan, Jeff},
  year = 2024,
  pages = {15787--15803}
}

@inproceedings{jaradehQuestionAnsweringScholarly2020,
  title = {Question Answering on Scholarly Knowledge Graphs},
  booktitle = {Digital Libraries for Open Knowledge},
  author = {Jaradeh, Mohamad Yaser and Stocker, Markus and Auer, S{\"o}ren},
  editor = {Hall, Mark and Mer{\v c}un, Tanja and Risse, Thomas and Duchateau, Fabien},
  year = 2020,
  pages = {19--32},
  publisher = {Springer International Publishing},
  address = {Cham},
  isbn = {978-3-030-54956-5}
}

@inproceedings{jiang2023structure,
  title = {A Structure and Content Prompt-Based Method for Knowledge Graph Question Answering over Scholarly Data.},
  booktitle = {{{QALD}}/{{SemREC}}@ {{ISWC}}},
  author = {Jiang, Longquan and Yan, Xi and Usbeck, Ricardo},
  year = 2023
}

@inproceedings{jiang2025kg,
  title = {Kg-Agent: {{An}} Efficient Autonomous Agent Framework for Complex Reasoning over Knowledge Graph},
  booktitle = {Proceedings of the 63rd Annual Meeting of the Association for Computational Linguistics (Volume 1: {{Long}} Papers)},
  author = {Jiang, Jinhao and Zhou, Kun and Zhao, Wayne Xin and Song, Yang and Zhu, Chen and Zhu, Hengshu and Wen, Ji-Rong},
  year = 2025,
  pages = {9505--9523}
}

@inproceedings{karkiSmallerLargeLanguage2025,
  title = {Smaller Large Language Models for Text-to-Sql: {{Performance}} Analysis and Optimal Performance},
  booktitle = {2025 International Conference on Inventive Computation Technologies ({{ICICT}})},
  author = {Karki, Sujan and Karki, Pukar and Shrestha, Binay Lal and Jha, Tantra Nath},
  year = 2025,
  pages = {1--7},
  doi = {10.1109/ICICT64420.2025.11005216}
}

@inproceedings{liu2024dora,
  title = {Dora: {{Weight-decomposed}} Low-Rank Adaptation},
  booktitle = {Forty-First International Conference on Machine Learning},
  author = {Liu, Shih-Yang and Wang, Chien-Yi and Yin, Hongxu and Molchanov, Pavlo and Wang, Yu-Chiang Frank and Cheng, Kwang-Ting and Chen, Min-Hung},
  year = 2024
}

@article{meloniExploringLargeLanguage2025,
  title = {Exploring Large Language Models for Scientific Question Answering via Natural Language to {{SPARQL}} Translation},
  author = {Meloni, Antonello and Reforgiato Recupero, Diego and Osborne, Francesco and {salatino}, angelo and Motta, Enrico and Vahadati, Sahar and Lehmann, Jens},
  year = 2025,
  month = aug,
  journal = {ACM Transactions on Intelligent Systems and Technology},
  publisher = {Association for Computing Machinery},
  address = {New York, NY, USA},
  issn = {2157-6904},
  doi = {10.1145/3757923}
}

@article{NGUYEN2025100135,
  title = {Fine-Tuning Text-to-{{SQL}} Models with Reinforcement-Learning Training Objectives},
  author = {Nguyen, Xuan-Bang and Phan, Xuan-Hieu and Piccardi, Massimo},
  year = 2025,
  journal = {Natural Language Processing Journal},
  volume = {10},
  pages = {100135},
  issn = {2949-7191},
  doi = {10.1016/j.nlp.2025.100135}
}

@inproceedings{perevalov2022qald,
  title = {Qald-9-plus: {{A}} Multilingual Dataset for Question Answering over Dbpedia and Wikidata Translated by Native Speakers},
  booktitle = {2022 {{IEEE}} 16th International Conference on Semantic Computing ({{ICSC}})},
  author = {Perevalov, Aleksandr and Diefenbach, Dennis and Usbeck, Ricardo and Both, Andreas},
  year = 2022,
  pages = {229--234},
  publisher = {IEEE}
}

@inproceedings{pourreza2025reasoningsql,
  title = {Reasoning-{{SQL}}: {{Reinforcement}} Learning with {{SQL}} Tailored Partial Rewards for Reasoning-Enhanced Text-to-{{SQL}}},
  booktitle = {Second Conference on Language Modeling},
  author = {Pourreza, Mohammadreza and Talaei, Shayan and Sun, Ruoxi and Wan, Xingchen and Li, Hailong and Mirhoseini, Azalia and Saberi, Amin and Arik, Sercan O},
  year = 2025
}

@article{qiEnhancingSPARQLQuery2024,
  title = {Enhancing {{SPARQL}} Query Generation for Knowledge Base Question Answering Systems by Learning to Correct Triplets},
  author = {Qi, Jiexing and Su, Chang and Guo, Zhixin and Wu, Lyuwen and Shen, Zanwei and Fu, Luoyi and Wang, Xinbing and Zhou, Chenghu},
  year = 2024,
  journal = {Applied Sciences},
  volume = {14},
  number = {1521},
  issn = {2076-3417},
  doi = {10.3390/app14041521}
}

@article{shao2025spurious,
  title = {Spurious Rewards: {{Rethinking}} Training Signals in Rlvr},
  author = {Shao, Rulin and Li, Shuyue Stella and Xin, Rui and Geng, Scott and Wang, Yiping and Oh, Sewoong and Du, Simon Shaolei and Lambert, Nathan and Min, Sewon and Krishna, Ranjay and others},
  year = 2025,
  journal = {arXiv preprint arXiv:2506.10947},
  eprint = {2506.10947},
  archiveprefix = {arXiv}
}

@misc{shaoDeepSeekMathPushingLimits2024,
  title = {{{DeepSeekMath}}: {{Pushing}} the {{Limits}} of {{Mathematical Reasoning}} in {{Open Language Models}}},
  shorttitle = {{{DeepSeekMath}}},
  author = {Shao, Zhihong and Wang, Peiyi and Zhu, Qihao and Xu, Runxin and Song, Junxiao and Bi, Xiao and Zhang, Haowei and Zhang, Mingchuan and Li, Y. K. and Wu, Y. and Guo, Daya},
  year = 2024,
  month = apr,
  number = {arXiv:2402.03300},
  eprint = {2402.03300},
  primaryclass = {cs},
  publisher = {arXiv},
  doi = {10.48550/arXiv.2402.03300},
  urldate = {2025-05-22},
  archiveprefix = {arXiv}
}

@misc{songEfficientTransferableAgentic2025a,
  title = {Efficient and {{Transferable Agentic Knowledge Graph RAG}} via {{Reinforcement Learning}}},
  author = {Song, Jinyeop and Wang, Song and Shun, Julian and Zhu, Yada},
  year = 2025,
  month = oct,
  number = {arXiv:2509.26383},
  eprint = {2509.26383},
  primaryclass = {cs},
  publisher = {arXiv},
  doi = {10.48550/arXiv.2509.26383},
  urldate = {2025-11-05},
  archiveprefix = {arXiv}
}

@inproceedings{stoisser2025sparks,
  title = {Sparks of Tabular Reasoning via {{Text2SQL}} Reinforcement Learning},
  booktitle = {The 4th Table Representation Learning Workshop at {{ACL}} 2025},
  author = {Stoisser, Josefa Lia and Martell, Marc Boubnovski and Fauqueur, Julien},
  year = 2025
}

@inproceedings{strappazon2025instruct,
  title = {Instruct-to-{{SPARQL}}: A Text-to-{{SPARQL}} Dataset for Training Wikidata Agents},
  booktitle = {{{ACM SIGIR}} Conference on Human Information Interaction and Retrieval},
  author = {Strappazon, Alexis and Granitzer, Michael and {Egyed-Zsigmond}, El{\H o}d and Mitrovic, Jelena and Amor, Mehdi Ben},
  year = 2025,
  publisher = {ACM}
}

@inproceedings{trivedi2017lc,
  title = {Lc-Quad: {{A}} Corpus for Complex Question Answering over Knowledge Graphs},
  booktitle = {International Semantic Web Conference},
  author = {Trivedi, Priyansh and Maheshwari, Gaurav and Dubey, Mohnish and Lehmann, Jens},
  year = 2017,
  pages = {210--218},
  publisher = {Springer}
}

@inproceedings{vossebeldLearningRefineAgentic2025,
  title = {Learning to {{Refine}}: {{An Agentic RL Approach}} for {{Iterative SPARQL Query Construction}}},
  shorttitle = {Learning to {{Refine}}},
  booktitle = {Proceedings of the {{Second International Workshop}} on {{Retrieval-Augmented Generation Enabled}} by {{Knowledge Graphs}} ({{RAGE-KG}} 2025)},
  author = {Vossebeld, Floris and Wang, Shenghui},
  editor = {Dobriy, Daniil and Tiwari, Sanju and D'Souza, Jennifer and Mihindukulasooriya, Nandana and Osborne, Francesco},
  year = 2025,
  month = nov,
  series = {{{CEUR Workshop Proceedings}}},
  volume = {4079},
  pages = {19--32},
  publisher = {CEUR},
  address = {Nara, Japan},
  issn = {1613-0073},
  urldate = {2025-12-02},
  langid = {english}
}

@article{wang2025octothinker,
  title = {Octothinker: {{Mid-training}} Incentivizes Reinforcement Learning Scaling},
  author = {Wang, Zengzhi and Zhou, Fan and Li, Xuefeng and Liu, Pengfei},
  year = 2025,
  journal = {arXiv preprint arXiv:2506.20512},
  eprint = {2506.20512},
  archiveprefix = {arXiv}
}

@misc{yang2025qwen3technicalreport,
  title = {Qwen3 Technical Report},
  author = {Yang, An and Li, Anfeng and Yang, Baosong and Zhang, Beichen and Hui, Binyuan and Zheng, Bo and Yu, Bowen and Gao, Chang and Huang, Chengen and Lv, Chenxu and Zheng, Chujie and Liu, Dayiheng and Zhou, Fan and Huang, Fei and Hu, Feng and Ge, Hao and Wei, Haoran and Lin, Huan and Tang, Jialong and Yang, Jian and Tu, Jianhong and Zhang, Jianwei and Yang, Jianxin and Yang, Jiaxi and Zhou, Jing and Zhou, Jingren and Lin, Junyang and Dang, Kai and Bao, Keqin and Yang, Kexin and Yu, Le and Deng, Lianghao and Li, Mei and Xue, Mingfeng and Li, Mingze and Zhang, Pei and Wang, Peng and Zhu, Qin and Men, Rui and Gao, Ruize and Liu, Shixuan and Luo, Shuang and Li, Tianhao and Tang, Tianyi and Yin, Wenbiao and Ren, Xingzhang and Wang, Xinyu and Zhang, Xinyu and Ren, Xuancheng and Fan, Yang and Su, Yang and Zhang, Yichang and Zhang, Yinger and Wan, Yu and Liu, Yuqiong and Wang, Zekun and Cui, Zeyu and Zhang, Zhenru and Zhou, Zhipeng and Qiu, Zihan},
  year = 2025,
  eprint = {2505.09388},
  primaryclass = {cs.CL},
  archiveprefix = {arXiv}
}

@inproceedings{zhang-zhao-2025-collaborative,
  title = {A Collaborative Reasoning Framework Powered by Reinforcement Learning and Large Language Models for Complex Questions Answering over Knowledge Graph},
  booktitle = {Proceedings of the 31st International Conference on Computational Linguistics},
  author = {Zhang, Zhiqiang and Zhao, Wen},
  editor = {Rambow, Owen and Wanner, Leo and Apidianaki, Marianna and {Al-Khalifa}, Hend and Eugenio, Barbara Di and Schockaert, Steven},
  year = 2025,
  month = jan,
  pages = {10672--10684},
  publisher = {Association for Computational Linguistics},
  address = {Abu Dhabi, UAE}
}

\appendix
\newpage
\section{Supplementary Material}
\label{app}

\subsection{Prompt Templates}
\label{app:prompts}
\subsubsection{System Prompt}
\label{app:system_prompt}
\begin{lstlisting}[]
Your task is to generate a syntactically and semantically correct SPARQL query that answers a given natural language question, using only the provided entities and relations.

Please follow these instructions:
- carefully analyze the given question, pay special attention to negations (not, etc.).
- Think step by step, reasoning through the transformation from question to query.
- Enclose your detailed reasoning in <think> ... </think> tags.
- Output the final SPARQL query - without any extra explanation or formatting.

Query Generation Rules:
- Only use the provided entities and relations; do not invent or infer additional ones.
- Use all provided entities and relations in the query.
- Do not use prefixes; write all URIs in full.
- By default, use SELECT DISTINCT in your queries, unless the context clearly requires otherwise.
- For yes/no questions, always use the ASK keyword to obtain a boolean result
- Carefully consider whether the answer should be the subject or object in each relevant triple pattern.
- Always use single quotes for literals

Example output format:
<think> Step-by-step reasoning here. </think> SPARQL query here
\end{lstlisting}

\subsubsection{User Prompt}
\label{app:user_prompt}
\begin{verbatim}
Generate a SPARQL query to answer the following question.

Question: {question}
Relevant Entities: {entities}
Relevant Relations: {relations}
\end{verbatim}

\subsection{Hyperparameters}

\subsubsection{Decoding Parameters}
\label{app:hyper_gen}

\begin{table}[H]
\floatconts
  {tab:gen-hparams}
  {\caption{Generation hyperparameters used for inference.}}
  {%
  \begin{tabular}{ll}
    \toprule
    \bfseries Parameter & \bfseries Value \\
    \midrule
    Base model           & Qwen3-1.7B (chat) \\
    Max input length     & 668 tokens \\
    Max new tokens       & 1024 tokens \\
    Temperature          & 0.6 \\
    Top-$p$              & 0.95 \\
    Min-$p$              & 0.0 \\
    Top-$k$              & 20 \\
    Beam search          & disabled \\
    Repetition penalty   & 1.0 \\
    Padding side         & left \\
    \bottomrule
  \end{tabular}
  }
\end{table}

\subsubsection{GRPO Training Parameters}
\label{app:hyper_grpo}
\begin{table}[H]
\floatconts
  {tab:grpo-hparams}
  {\caption{Hyperparameters for GRPO training.}}
  {%
  \begin{tabular}{ll}
    \toprule
    \bfseries Parameter & \bfseries Value \\
    \midrule
    Base model                    & Qwen3-1.7B (chat) \\
    Training set size             & 7000 \\
    Batch size (per device)       & 4 \\
    Gradient Accumulation Steps   & 16 \\
    Group size                    & 4 \\
    Max new tokens                & 1024 \\
    Learning rate                 & 1e-6 \\
    Optimizer                     & AdamW \\
    Beta 1 / Beta 2               & 0.9 / 0.999 \\
    Weight decay                  & 0.0 \\
    LR schedule                   & linear \\
    Advantage normalization       & per-group \\
    KL penalty ($\beta$)          & 0.04 \\
    KL target policy              & initial pretrained model \\
    Clipping ($\varepsilon$) & 0.2 \\
    Mixed precision               & bfloat16 \\
    Random seed                   & 42 \\
    \bottomrule
  \end{tabular}
  }
\end{table}

\subsubsection{DoRA Fine-tuning Parameters}
\label{app:hyper_dora}

\begin{table}[H]
\floatconts
  {tab:dora-hparams}
  {\caption{Hyperparameters for DoRA supervised fine-tuning.}}
  {%
  \begin{tabular}{ll}
    \toprule
    \bfseries Parameter & \bfseries Value \\
    \midrule
    Base model           & Qwen3-1.7B (chat) \\
    Training epochs      & 1 \\
    Batch size (per GPU) & 8 \\
    Gradient accumulation& 8 \\
    Effective batch size & 64 \\
    Learning rate        & 2e-5 \\
    Optimizer            & AdamW \\
    Weight decay         & 0.0 \\
    LR schedule          & linear \\
    Warmup ratio         & 0.0 \\
    Max sequence length  & 1024 tokens \\
    Loss                 & Cross-entropy on target SPARQL \\
    DoRA rank            & 8 \\
    DoRA alpha           & 32 \\
    DoRA dropout         & 0.05 \\
    Target modules       & q, k, v, o projections \\
    Mixed precision      & bfloat16 \\
    \bottomrule
  \end{tabular}
  }
\end{table}

%\subsection{GRPO Error Patterns}
%\label{app:error_patterns}
%\begin{figure}[H]
%    \centering
%    \begin{tikzpicture}
%        \begin{axis}[
%            ybar,
%            ymin=0, ymax=40,
%            bar width=10pt,
%            width=0.9\textwidth,
%            height=6cm,
%            enlarge x limits=0.1,
%            ylabel={Share of GRPO failures (\%)},
%            symbolic x coords={
%                Disjunction,
%                Negation,
%                Aggregation,
%                Multi-intent,
%                Temporal,
%                Syntax/Schema
%            },
%            xtick=data,
%            xticklabel style={rotate=30, anchor=east},
%            ymajorgrids=true,
%            grid style={dashed,very thin},
%        ]
%            % GRPO error shares per category (100-sample)
%            \addplot coordinates {
%                (Disjunction,20)
%                (Negation,24)
%                (Aggregation,21)
%                (Multi-intent,32)
%                (Temporal,23)
%                (Syntax/Schema,33)
%            };
%        \end{axis}
%    \end{tikzpicture}
%    \caption[Distribution of error categories in a random sample]{
%        Distribution of error categories in a random sample of 100 incorrect GRPO generations.
%        Bars show the proportion of failures in which each category is present.
%        Categories are not mutually exclusive.
%    }
%    \label{fig:grpo_error_categories}
%\end{figure}

\end{document}